\PassOptionsToPackage{table}{xcolor}

\documentclass{article} 
\usepackage{iclr2024_conference,times}

\usepackage{xcolor}
\usepackage{url}
\usepackage{latexsym}
\usepackage{graphicx}
\usepackage{amsmath}
\usepackage{amssymb}
\usepackage{booktabs}
\usepackage{multirow}

\definecolor{nicergreen}{rgb}{0.13, 0.54, 0.13}
\definecolor{nicered}{rgb}{0.83, 0.16, 0.16}
\newcommand\showdiff[1]{\textbf{\textcolor{nicergreen}{#1}}}
\newcommand\showdiffn[1]{\textbf{\textcolor{nicered}{#1}}}

\usepackage{xspace}
\usepackage[T1]{fontenc}
\usepackage{multirow}
\usepackage{tabularx}

\usepackage{wrapfig}
\usepackage{tabulary}

\usepackage[utf8]{inputenc}
\newcommand{\nop}[1]{}
\usepackage{listings}
\usepackage{caption}
\usepackage{subcaption}
\usepackage{hyperref}

\newcommand{\NAME}{PaLI-17B\xspace}
\newcommand{\NEWNAME}{PaLI-3\xspace}

\newcommand{\prompt}[1]{``\emph{#1}''}
\newcommand{\exid}{$\langle\mathrm{extra\_id\_0}\rangle$\xspace}

\newcommand{\token}[2]{$\langle\mathrm{#1}\_{#2}\rangle$\xspace}

\newcommand{\lang}{$\langle\mathrm{lang}\rangle$\xspace}
\newcommand{\pos}{$\langle\mathrm{pos}\rangle$\xspace}

\newcommand{\imres}[1]{#1$\times$#1}

\newcommand{\sotamodel}[1]{\textcolor{gray}{{\footnotesize #1}}}

\title{PaLI-3 Vision Language Models:\\
Smaller, Faster, Stronger}
\author{\vspace{0.2em} Xi Chen$_1^{\star}$\quad Xiao Wang$_2^{\star}$\quad Lucas Beyer$_2^{\star}$ \quad Alexander Kolesnikov$_2$ \quad Jialin Wu$_1$ \\ 
\vspace{0.2em} \bf{Paul Voigtlaender$_1$ \quad Basil Mustafa$_2$ 
\quad Sebastian Goodman$_1$ \quad Ibrahim Alabdulmohsin$_2$} \\
\vspace{0.2em} \bf{Piotr Padlewski$_2$ \quad Daniel Salz$_1$ \quad Xi Xiong$_3$ \quad Daniel Vlasic$_3$ \quad Filip Pavetic$_2$} \\
\vspace{0.2em} \bf{Keran Rong$_2$ \quad Tianli Yu$_3$ \quad Daniel Keysers$_2$ \quad Xiaohua Zhai$_2^{\star\dagger}$ \quad Radu Soricut$_1^{\dagger}$\vspace{0.4em}} \\
$^{1}$Google Research\quad $^{2}$Google DeepMind \quad $^{3}$Google Cloud
}

\begin{document}

\maketitle
{\let\thefootnote\relax\footnote{
{\hspace{-2em}$^{\star}$Core contributors, $^{\dagger}$Project leads}
}
}

\begin{abstract}
This paper presents \NEWNAME, a smaller, faster, and stronger vision language model (VLM) that compares favorably to similar models that are 10x larger.
As part of arriving at this strong performance, we compare Vision Transformer (ViT) models pretrained using classification objectives to contrastively (SigLIP) pretrained ones. 
We find that, while slightly underperforming on standard image classification benchmarks, SigLIP-based PaLI shows superior performance across various multimodal benchmarks, especially on localization and visually-situated text understanding.
We scale the SigLIP image encoder up to 2 billion parameters, and achieves a new state-of-the-art on multilingual cross-modal retrieval.
We hope that \NEWNAME, at only 5B parameters, rekindles research on fundamental pieces of complex VLMs, and could fuel a new generation of scaled-up models.
\end{abstract}

\section{Introduction}\label{sec:intro}

The scaling of vision-language models (VLM) to tens and even hundreds of billions of parameters~\citep{pali2, alayrac2022flamingo, chen2023palix,  palme} has shown ever-increasing performance. Meanwhile, models at a smaller scale remain critical, as they are more practical to train and serve, more environmentally-friendly, and support faster research cycles for model design.

In the spirit of focusing on smaller-scale modeling, we present \NEWNAME, the third-generation family of PaLI~\citep{pali2} models.
Using a pretrained backbone with only 5B total parameters, we refine the training recipe and achieve competitive and new state-of-the-art (SOTA) results on various VLM benchmarks.
Our new recipe has three main components: contrastive pretraining of image encoder on web-scale image-text data~\citep{siglip}, an improved dataset mixture for PaLI multimodal training, and training at higher resolutions. 

\NEWNAME achieves new SOTA results on tasks that require visually-situated text understanding and object localization, including eight visually-situated text understanding tasks and the referring expression segmentation task on RefCOCO~\citep{refcoco}, along with strong performance on a wide range of classical vision tasks. As part of this work, we also introduce a SOTA multilingual contrastive vision model scaled to 2B parameters, obtained using the recently-introduced SigLIP recipe~\citep{siglip}. Additionally, we perform focused ablation studies to compare the classification pretrained Vision Transformer (ViT) backbones~\citep{vit} with contrastively pretrained ones (SigLIP). This further confirms the viability of pretraining visual encoders on noisy web-scale image-text data, as a preferable alternative to training on classification-style data.

Our contributions are summarized as follows:
 \begin{enumerate}
     \item We compare classification pretrained ViT models~\citep{vit} to contrastively pretrained SigLIP models using the PaLI framework~\citep{pali2}. We find that the contrastively pretrained models work significantly better for visually-situated text understanding tasks and localization tasks.
     \item Our model achieves SOTA performance on 10+ diverse vision-language benchmarks while being 10x smaller in size compared to the current SOTA model~\cite{chen2023palix}. For understanding visually-situated text, the improvement is by a particularly large margin.
     \item Despite not pretraining on any video data, our model achieves new SOTA on several video QA benchmarks, indicative of powerful generalization abilities.
     \item We introduce the 2 billion parameter (ViT-G) multilingual SigLIP model trained on WebLI~\citep{pali2}, which sets a new state-of-the-art on the multilingual cross-modal retrieval benchmark~\cite{Thapliyal2022Crossmodal3600AM} across 36 languages.
 \end{enumerate}

\section{Related Work}\label{sec:related_work}

Recent large vision language models (VLMs) use pretrained image encoders as part of the larger model, some pretrain it with supervised classification (PaLI~\citep{pali2}, PaLI-X~\citep{chen2023palix}, PaLM-E~\citep{palme}), some use pretrained CLIP encoders (BLIPv2~\citep{blipv2}, CrossTVR~\citep{dai2023finegrained}, ChatBridge~\citep{zhao2023chatbridge}) and some with custom multimodal pretraining (Flamingo~\citep{alayrac2022flamingo}, BEiT3~\citep{beitv3}, CoCa~\citep{yu2022coca}, SimVLM~\citep{wang2021simvlm}).
In this paper we compare two dominant ways to pretrain image encoders using the PaLI framework: classification pretraining using large weakly labeled datasets \citep[JFT, as in][]{kolesnikov2020bit,zhai2022scaling,vit-22b} and contrastive pretraining on web-scale noisy data \citep[WebLI, as in][]{siglip}.

A recent finding spanning across PaLI~\citep{pali2} and PaLI-X~\citep{chen2023palix} is that scaling up the classification pretrained image encoder seems more promising than was previously believed \citep{alayrac2022flamingo}.
Specifically, while classic image-only benchmarks such as ImageNet seem to indicate saturating performance from scaling pretraining of image-only models~\citep{zhai2022scaling}, PaLI shows that by scaling up the vision encoder from ViT-G (2B) to ViT-e (4B), the improvements on VL tasks are more noticeable than on ImageNet.
PaLI-X further scaled up both the vision and language components, showing that these larger image encoders keep bringing benefit when plugged into large VLMs.
This finding suggests that there is more to be found regarding the pretraining of image encoders in the context of VLMs, which may lead to different conclusions when looking at VLM tasks as compared to of ``pure'' vision tasks.
In this paper, we dive into the impact of the image encoder for VLMs, by directly comparing classification pretrained vision models to contrastively pretrained ones, and reveal that the latter are vastly superior on a variety of tasks, especially localization and visually-situated text understanding.

One can split multimodal understanding capabilities into largely two categories: natural scene understanding (captioning, VQA, object detection/localization), and visually-situated text understanding (document and infographics QA).
These groups of tasks require different granularity of understanding, and previous VLMs have largely focused on one type of tasks, leading to their training recipes being attuned to that type of tasks. For example PaLI-17B~\citep{pali2} and Pix2struct~\citep{lee2022pix2struct} showed strong performance only on one of the two categories, respectively. The recent PaLI-X~\citep{chen2023palix} achieves SOTA performance on both categories, based on an improved OCR-related training recipe, and a significantly larger 55B parameter model. In this work, we combine the advantage of contrastively-pretrained ViT and a further improved and balanced training recipe into \NEWNAME, and demonstrate that SOTA level performance on both the above categories of multimodal understanding is achievable even at 5B parameter scale.

\section{Model}\label{sec:model}

\begin{figure}[t]
    \centering
    \includegraphics[width=\linewidth]{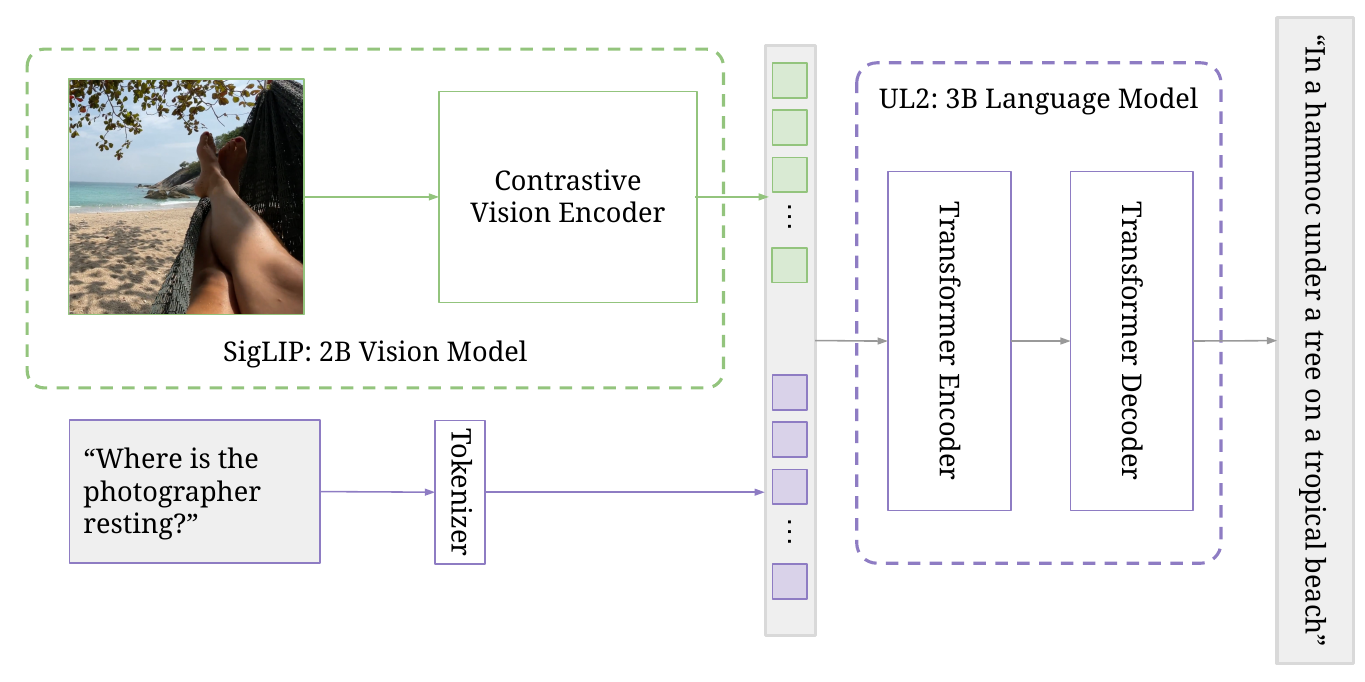}
    \caption{Overview of the \NEWNAME (5B) model: images are encoded into visual tokens individually by the contrastively pretrained 2B SigLIP vision model.
    Along with a query, these visual tokens are passed to an 3B encoder-decoder UL2 Transformer which produces the desired answer.
    In such a setup, a contrastively pretrained model provides significantly more useful tokens than one classification pretrained model as in previous PaLI models.}
    \label{fig:sketch}
\end{figure}

\subsection{Architecture}\label{section:arch}

On a high level, the architecture follows \citet{pali2,chen2023palix}: a ViT encodes the image into tokens which, together with text input (the question, prompt, instruction), are passed to an encoder-decoder transformer \citep{vasvani2017attention} that generates a text output.

\paragraph{Visual component}
The vision backbone of \NEWNAME is initialized from a contrastively pretrained ViT-G/14\footnote{The embedding dimension was changed from 1664 to 1536 for better hardware utilization.} \citep{zhai2022scaling} model (approx. 2B parameters) using the SigLIP \citep{siglip} training recipe. 
In short, an image embedding ViT-G/14 and a text embedding transformer are trained to separately embed images and texts, such that a binary classifier using the sigmoid cross-entropy of the dot product of image and text embeddings correctly classifies whether the respective image and text correspond to each other or not.
This is similar to CLIP \citep{radford2021clip} and ALIGN \citep{jia2021align}, but was shown to be more efficient, scalable, and robust \citep{siglip}.
This is done in order to pretrain the ViT image embedding component, hence the text embedding transformer is discarded when inserting the ViT into PaLI.

\paragraph{Full PaLI model}
The outputs of the ViT image encoder before pooling form the visual tokens, which are linearly projected and prepended to the embedded input text tokens.
Together, these tokens are passed into a pretrained 3B parameter UL2 encoder-decoder language model \citep{tay2023ul2}, which generates text output.
The text input to the model typically consists of a prompt that describes the type of task (e.g., \prompt{Generate the alt\_text in \lang at \pos} for captioning tasks) and encode necessary textual input for the task (e.g., \prompt{Answer in \lang: \{question\} } for VQA tasks).

\subsection{Stages of Training}
The training procedure is similar to that of PaLI and PaLI-X and consists of multiple stages:

\paragraph{Stage 0: Unimodal pretraining.} The image encoder is pretrained contrastively on image-text pairs from the web, following the SigLIP training protocol \citep{siglip}. This differs from PaLI and PaLI-X, where a JFT classification pretrained encoder was used.
We use a model-based filtering approach similar to that in \citet{schuhmann2021laion} that retains about 40\% of the pairs.
The image encoder is trained at resolution \imres{224}.
The text encoder-decoder is a 3B UL2 model trained following the mixture of denoisers procedure described by \citet{tay2023ul2}.

\paragraph{Stage 1: Multimodal training.} Here, the image encoder is combined with the text encoder-decoder as described earlier and in Figure~\ref{fig:sketch}.
Then, this combined PaLI model is trained on a multimodal task and data mixture, albeit keeping the image encoder \texttt{frozen}  and using its native (\imres{224}) resolution.
The main mixture component is again derived from the WebLI dataset by heuristic filtering of the text quality and using the SplitCap training objective \citep{pali2}.
Further ingredients inherited from \citep{pali2} are multilingual captioning on CC3M-35L and WebLI OCR, cross-lingual VQA and VQG using VQ$^2$A-CC3M-35L, object-aware VQA, as well as object detection.
Notably, we do not include task or data derived from video (this was done in PaLI-X), but \NEWNAME retains competitive performance on these benchmarks thanks to its strong image encoder. 
We do, however, further improve document and text understanding capabilities by enriching WebLI with PDF documents with dense text and web-images described as posters or documents, in over 100 languages.

\paragraph{Stage 2: Resolution increase.} High-resolution input is a widely accepted way of increasing performance, both due to making more details from the image perceptible, and due to increasing model power via increased sequence length.
We increase \NEWNAME's resolution by fine-tuning the whole model (unfreezing the image encoder) with a short curriculum of increasing resolutions, keeping checkpoints at \imres{812} and \imres{1064} resolution. The data mixture focuses on the part that involves visually-situated text and object detection.

\paragraph{Task specialization (transfer).} Finally, for each individual task (benchmark), we fine-tune the \NEWNAME model with frozen ViT image encoder on the task's training data as described in the corresponding section.
For most tasks, we fine-tune the \imres{812} resolution checkpoint, but for two document understanding tasks, we go up to \imres{1064} resolution.

\section{Experiments}\label{sec:exp}

\subsection{Classification or Contrastively pretrained ViT?}
\begin{table}[t]
  \newcolumntype{C}{>{\centering\arraybackslash}X}
  \newcolumntype{R}{>{\raggedleft\arraybackslash}X}
  \setlength{\tabcolsep}{0pt}
  \setlength{\extrarowheight}{5pt}
  \renewcommand{\arraystretch}{0.75}
  \centering
  \caption{Performance comparison between contrastively pre-trained (``SigLIP'') models and classification pre-trained (``Classif'') ViT image encoders using the same PaLI setup, across a wide range of tasks.
  While linear classification few-shot probing (first column) suggests SigLIP encoders are worse across many tasks, when plugged into PaLI and transferred, they show clear improvements.
  On the most complicated and detailed image understanding tasks, SigLIP models outperform Classif models by a large margin.
  Captioning numbers are CIDEr scores, where XM3600 shows the English performance in the first column, and the average across other languages in the second column. RefCOCO numbers are mIoU scores (details in Section \ref{sec:exp:refseg}).
  }\label{table:controlled_transfers}
  \begin{tabularx}{\linewidth}{
  p{0.5cm}p{1.2cm}p{0.1cm}
  Cp{0.2cm}Cp{0.1cm}CCp{0.2cm}CCCp{0.2cm}CCCp{0.2cm}}
    \toprule[1pt]
&&& \bf{Probe} && \multicolumn{4}{c}{\bf{Captioning}} && \multicolumn{3}{c}{\bf{VQA}} && \multicolumn{3}{c}{\bf{RefCOCO}} \\
     \cmidrule[0.5pt]{4-4} \cmidrule[0.5pt]{6-9} \cmidrule[0.5pt]{11-13} \cmidrule[0.5pt]{15-17} 
&&& 8 tasks      && COCO && \multicolumn{2}{c}{XM3600} && v2 & OK & Text && val & + & g \\
    \midrule

\multirow{2}{*}{\rotatebox{90}{\hspace*{-2pt}G/14}}
& Classif && 88.1 && 139.9 && 94.5 & 44.7 && 76.7 & 57.2 & 31.9 && 51.6 & 43.5 & 43.4  \\
& SigLIP && \showdiffn{-2.5} && \showdiff{+0.4} && \showdiff{+1.6} & \showdiff{+0.7} && \showdiff{+0.8} & \showdiff{+1.4} & \showdiff{+18.7} && \showdiff{+15.1} & \showdiff{+19.1} & \showdiff{+17.7}  \\

\midrule

\multirow{2}{*}{\rotatebox{90}{\hspace*{-2pt}L/16}}
& Classif && 86.2 && 132.6 && 93.0 & 42.3 && 73.7 & 55.6 & 24.9 && 46.9 & 38.8 & 38.8 \\
& SigLIP && \showdiffn{-2.8} && \showdiff{+3.2} && \showdiff{+1.4} & \showdiff{+1.4} && \showdiff{+1.9} & \showdiff{+1.9} & \showdiff{+16.2} && \showdiff{+17.4} & \showdiff{+20.9} & \showdiff{+20.1}  \\

\midrule

\multirow{2}{*}{\rotatebox{90}{\hspace*{-2pt}B/16}}
& Classif && 83.7 && 127.7 && 91.7 & 40.7 && 72.3 & 54.7 & 22.5 && 46.3 & 38.1 & 38.4 \\
& SigLIP && \showdiffn{-2.6} && \showdiff{+3.6} && \showdiffn{-2.0} & \showdiffn{-0.2} && \showdiff{+1.4} & \showdiff{+0.9} & \showdiff{+13.3} && \showdiff{+16.8} & \showdiff{+19.6} & \showdiff{+19.3}  \\
    \bottomrule[1pt]
  \end{tabularx}
\end{table}

We first perform a controlled comparison of different ViT models within the PaLI framework. 
We consider two types of ViT models: classification pretrained (``Classif'') on the JFT dataset and contrastively pretrained on the WebLI dataset (``SigLIP'').
We perform these experiments using a fixed \imres{224} resolution (i.e.\ only include Stage 1) to save compute.
We further shorten the Stage 1 phase to 20\% of the full \NEWNAME schedule used in the remainder of this paper.

The results in Table~\ref{table:controlled_transfers} paint a clear picture overall:
While the few-shot linear classification \citep{vit} of SigLIP models falls behind, when used in PaLI-3, SigLIP models provide moderate gains on ``simpler'' tasks such as captioning and question-answering, and large gains for more ``complicated'' scene-text and spatial understanding tasks such as TextVQA and RefCOCO variants.
This motivates the departure from classification pretrained image encoders, and switching to sigmoid-contrastively pretrained ones for building \NEWNAME.

\subsection{Visually-situated Text Understanding} \label{sec:exp:text_understanding}

We evaluate \NEWNAME on visually-situated text understanding tasks: TextCaps~\citep{textcaps}, TextVQA~\citep{textvqa}, STVQA~\citep{stvqa}, OCRVQA~\citep{ocrvqa}, InfographicVQA~\citep{mathew2022infographicvqa}, DocVQA~\citep{mathew2021docvqa}, ChartQA~\citep{masry2022chartqa}, Scree2Words~\citep{screen2words}, and WidgetCap~\citep{li-etal-2020-widget}. The images in those datasets span a wide range of domains such as natural images, illustrations, documents and user interfaces.

For the InfographicVQA and DocVQA benchmarks we fine-tune the \imres{1064} resolution model, all others use the \imres{812} one.
We report the standard metrics for each benchmark, namely: CIDEr score for all the Image Captioning benchmarks; VQA accuracy for VQAv2, OKVQA, and TextVQA; Average Normalized Levenshtein Similarity (ANLS) for ST-VQA, DocVQA and InfographicsVQA; Exact match (EM) for TallyQA, OCR-VQA and AI2D; Relaxed-Accuracy (RA) for ChartQA.
For visually-situated text understanding, external OCR systems are usually leveraged to provide OCR annotations of the image as and additional input to the model for boosting performance. Here we follow~\citep{chen2023palix} and report the result of finetuning \NEWNAME both with and without OCR inputs. The OCR annotations are obtained using the same service as that for the training set. As shown in Table~\ref{table:scene_text_like}, \NEWNAME achieves SOTA performance on a vast majority of the captioning and VQA benchmarks both with and without external OCR input. The exception is AI2D and ChartQA, which require not just understanding but also strong reasoning capability over diagrams and charts, respectively. For these two benchmarks, \NEWNAME falls slightly behind PaLI-X~\citep{chen2023palix} likely due to the latter's significantly larger 32B LLM being better at reasoning.

\begin{table*}
\centering
\caption{Results on benchmarks more focused on understanding visually-situated text. TextCaps, TextVQA, STVQA, InfographicVQA and DocVQA are all evaluated using the corresponding official evaluation server. Methods marked by "$^\dagger$" are trained on additional VQA data similar to the target benchmark, before finetuning on the target benchmark. For ChartQA, we compare with similar setups by finetuning without chain-of-thought or similar prompting techniques. The SOTA models are (a) \cite{chen2023palix}, (b) \cite{powalski2021going}, (c) \cite{peng2022ernie}.}
\label{table:scene_text_like}
\begin{tabular}{lc@{\hspace{0.20cm}}c@{\hspace{0.20cm}}c@{\hspace{0.20cm}}c@{\hspace{0.20cm}}c@{\hspace{0.20cm}}c@{\hspace{0.20cm}}c@{\hspace{0.20cm}}c@{\hspace{0.20cm}}c@{\hspace{0.20cm}}c@{\hspace{0.20cm}}|c@{\hspace{0.20cm}}}
\toprule
& Text & Text & ST & OCR & Info & Doc & \multirow{2}{*}{AI2D} & Chart & Screen2 & Widget & Avg. of\\
Model & Caps & VQA & VQA & VQA & VQA & VQA & & QA & Words & Cap & first 8\\
\midrule
\multicolumn{4}{l}{\textit{\textbf{with} OCR pipeline input}} &&&&&& \\
\midrule
\multirow{2}{*}{SOTA}
& 163.7 & \textbf{80.78} & 84.5 & 77.3 & 61.2 & 88.4 & \textbf{81.4} & \textbf{72.3} & \multirow{2}{*}{-} & \multirow{2}{*}{-} & \multirow{2}{*}{88.7} \\
&
\sotamodel{(a)} &
\sotamodel{(a)} &
\sotamodel{(a)} &
\sotamodel{(a)} &
\sotamodel{(b)$^\dagger$} &
\sotamodel{(c)$^\dagger$} &
\sotamodel{(a)} &
\sotamodel{(a)} & & \\
\NEWNAME & \textbf{164.3}  & \textbf{80.78}  & \textbf{85.7} & \textbf{77.8} & \textbf{62.4} & \textbf{88.6} & 75.2 & 69.5 & - & - & 88.0 (\textcolor{red}{-0.7}) \\
\midrule
\multicolumn{4}{l}{\textit{\textbf{without} OCR pipeline input}} \\
\midrule
\multirow{2}{*}{SOTA}
& 147.0 & 71.44 & 79.9 & 75.0 & 49.2 & 80.0 & \textbf{81.2} & \textbf{70.9} & 127.9 & 153.0 & \multirow{2}{*}{81.8} \\
&
\sotamodel{(a)} &
\sotamodel{(a)} &
\sotamodel{(a)} &
\sotamodel{(a)} &
\sotamodel{(a)} &
\sotamodel{(a)} &
\sotamodel{(a)} &
\sotamodel{(a)} &
\sotamodel{(a)} &
\sotamodel{(a)} \\
\NEWNAME & \textbf{158.8} & \textbf{79.51} & \textbf{84.1} & \textbf{76.7} & \textbf{57.8} & \textbf{87.6} & 75.2 & 70.0 & \textbf{130.7} & \textbf{159.8} & 86.2 (\textcolor{teal}{+4.4}) \\
\bottomrule
\end{tabular}
\end{table*}

Averaging over the 8 benchmarks in Table~\ref{table:scene_text_like} that have results in all settings, \NEWNAME is only 0.7 points behind all SOTA methods combined in the setting where external OCR systems are used. However, in the setting without such external system, \NEWNAME has a significant 4.4 point advantage over all SOTA methods combined. For TextCaps, TextVQA, InfographicVQA and DocVQA this advantage is 8 points or more. Finally, we can see that \NEWNAME without any external OCR system is only 1.8 points behind relying on one, suggesting the image encoder learns a strong intrinsic OCR capability. 

\subsection{Referring Expression Segmentation}\label{sec:exp:refseg}
We extend \NEWNAME with the capability to predict segmentation masks via language-like output.
To this end, we make use of the vector-quantized variational auto-encoder (VQ-VAE) from \citet{ning2023all}.
The VQ-VAE is trained to learn a discrete codebook of 128 mask tokens.
Its encoder can tokenize a $64\times64$ pixels segmentation mask into 16 mask tokens, which its decoder can convert back.
We train \NEWNAME to predict a single segmentation mask.
First, \NEWNAME outputs 4 coordinates as text, representing a bounding box.
This is followed by 16 mask tokens that represent the mask inside the bounding box.

We fine-tune \NEWNAME on the combined training sets of RefCOCO, RefCOCO+, and RefCOCOg \citep{refcoco}\footnote{We removed all validation and test images from the training set for both PaLI and the VQ-VAE} at \imres{812} resolution.
Each training example consists of a referring expression (e.g.\ ``the large brown dog on the left''), and a box with segmentation mask.
We prompt PaLI with the prompt \prompt{detect: the large brown dog on the left \exid}, and the target is a sequence like \prompt{348 543 684 664 \token{mask\_token}{81} \dots \token{mask\_token}{10}}.
The target sequence contains 16 mask tokens between $0$ and $127$ that are generated by the VQ-VAE encoder using the segmentation mask cropped and resized to $64\times64$ as input.

The results in Table~\ref{table:controlled_transfers} demonstrate that contrastive pretraining is much more effective than classification pretraining for localization task of this type. Table ~\ref{table:refexp_seg} shows that the full \NEWNAME model is able to slightly outperform the state of the art in referring expression segmentation.

\begin{table*}
\centering
\caption{PaLI referring expression segmentation results on RefCOCO~\citep{refcoco} variants. All results are mIoU on the {\tt val} split.}\label{table:refexp_seg}
\begin{tabular}{lccc}
\toprule 
Model & RefCOCO & RefCOCO+ & G-Ref \\
\midrule 
RefTr \citep{li2012reftr} & 74.34 & 66.75 & 66.63 \\
PolyFormer \citep{polyformer} & 76.94 & 72.15  & 71.15 \\
\NEWNAME (Ours) & \textbf{77.33} & \textbf{73.53} & \textbf{72.72} \\
\bottomrule
\end{tabular}
\end{table*}

\subsection{Natural Image Understanding}\label{sec:exp:finetuning}

In this section, we evaluate \NEWNAME on general vision-language understanding tasks, including COCO captions~\citep{cocokarp} and VQAv2~\citep{vqav2} which target general visual understanding, OKVQA~\citep{marino2019okvqa} which focuses on knowledge-based understanding and TallyQA~\citep{acharya2019tallyqa} which measures performance on counting under textual guidance.
All results for the benchmarks presented in Table~\ref{table:non_ocr} use \imres{812} resolution.
As in previous work, they employ no external OCR module, since these benchmarks rarely involve text in the image. 

Overall, \NEWNAME shows very strong performance on these benchmarks despite its significantly smaller size compared to recent SOTA models.
For COCO, \NEWNAME outperforms all models but BEiT-3 and the 17B and 55B PaLI. On VQAv2 and TallyQA, \NEWNAME exceeds all previous models except PaLI-X, with a less than 1 point gap on VQAv2. For the knowledge-intensive OKVQA task, which usually benefits from a large language component, \NEWNAME is only behind PaLM-E (562B) and PaLI-X (55B) but still outperforms the 32-shot Flamingo (80B) model.

\begin{table}[h]
\centering
\caption{Results on COCO Captions (Karpathy split), VQAv2, OKVQA, and TallyQA.
(*Flamingo reports 32 shot result). Underscored numbers indicate that \NEWNAME is only behind the 55B PaLI-X and is better than all other models in the list.}
\label{table:non_ocr}
\begin{tabular}{lccccccccc}
\toprule
& COCO && \multicolumn{2}{c}{VQAv2} && OKVQA && \multicolumn{2}{c}{TallyQA} \\
\cmidrule{2-2} \cmidrule{4-5} \cmidrule{7-7} \cmidrule{9-10}
Model & Karp.-test & & test-dev & test-std && val && Simple & Complex \\
\midrule
SimVLM & 143.3 && 80.03 & 80.34 && - && - & -\\
CoCa (2.1B) & 143.6 && 82.3 & 82.3 && - && - & -\\
GIT (0.7B) & 144.8 & & 78.56 & 78.81 && - && - & -\\
GIT2 (5.1B) & 145.0 && 81.74 & 81.92 && - && - & -\\
OFA (0.9B) & 145.3 && 82.0 & 82.0 && - && - & -\\
Flamingo (80B) & 138.1 && 82.0 & 82.1 && 57.8$^*$ && - & -\\
BEiT-3 (1.9B) & 147.6 && 84.2 & 84.0 && - && - & -\\
PaLM-E (562B) & 138.7 && 80.0 & - && \textbf{66.1} & - & - \\
MoVie & - && 69.26 & - && - && 74.9 & 56.8 \\
\NAME & 149.1 & & 84.3 & 84.3 && 64.5 && 81.7 & 70.9 \\
PaLI-X (55B) & \textbf{149.2} && \textbf{86.0} & \textbf{86.1} && \textbf{66.1} && \textbf{86.0} & \textbf{75.6} \\
\midrule
\NEWNAME (5B) & 145.9 && \underline{85.0} & \underline{85.2} && 60.1 && \underline{83.3} & 70.5\\
\bottomrule
\end{tabular}
\end{table}

\subsection{Video Captioning and Question Answering}\label{sec:exp:video}

We fine-tune and evaluate the \NEWNAME model on 4 video captioning benchmarks: MSR-VTT \citep{xu2016msr}, VATEX \citep{wang2019vatex}, ActivityNet Captions \citep{krishna2017dense}, and Spoken Moments in Time \citep{monfort2021spoken}.
We do the same for 3 video question-answering benchmarks: NExT-QA \citep{xiao2021next}, MSR-VTT-QA \citep{xu2017video}, and ActivityNet-QA \citep{yu2019activitynet}.
A brief description of each benchmark and its usage is provided in Appendix~\ref{appendix:video_results}.

Following the setup from PaLI-X~\citep{chen2023palix}, we fine-tune our model using the Stage 1 checkpoint with \imres{224} resolution for each task separately.
We sample at most 16 frames with a fixed temporal stride for each benchmark. Each frame is independently processed by the ViT image encoder, the resulting visual tokens are simply concatenated, leading to up to 4096 visual tokens. A key difference from the PaLI-X setup is that there is no video data in \NEWNAME pretraining, meaning \NEWNAME has never seen multi-frame inputs during pretraining.

Despite not being pretrained with video data, \NEWNAME achieves excellent video QA results with a small model size: a new state of the art performance on MSR-VTT-QA and ActivityNet-QA, and competitive results on NextQA. The consistent improvements on image and video QA highlight the benefits of adopting the contrastive ViTs. \NEWNAME also achieves respectable video captioning results, under-performing the SOTA by only 3 CIDEr points on average. Considering the model size, \NEWNAME appears to be an excellent choice in terms of both performance and practicality.

\begin{table}[t]
\centering
\caption{Results for Video Captioning and Video-QA using up to 16 frames. $\dagger$GIT2 directly optimizes the CIDEr metric.
mPLUG-2 is \citet{mplug2}, PaLI-X is \citet{chen2023palix}, GIT2 is \citet{wang2022git}, and Flamingo-32 is the 32-shot variant of \citet{alayrac2022flamingo}.
}
\label{table:video}
\begin{tabular}{lccccccc}
\toprule
& \multicolumn{2}{c}{MSR-VTT} & \multicolumn{2}{c}{Activity-Net} & VATEX & SMIT & NExT-QA \\
\cmidrule{2-8}
Method & Caption & QA & Caption & QA & Caption & Caption & QA \\
\midrule
Prior SOTA & \textbf{80.3} & 48.0 & \textbf{54.9} & 49.4 & 94.0$^\dagger$ & \textbf{43.5} &  \textbf{38.3} \\
& \sotamodel{mPLUG-2} &
\sotamodel{mPLUG-2} &
\sotamodel{PaLI-X} &
\sotamodel{PaLI-X} &
\sotamodel{GIT2} &
\sotamodel{PaLI-X} &
\sotamodel{Flamingo-32}
\\
\midrule
\NEWNAME & 78.3 & \textbf{49.3} & 50.8 & \textbf{51.2} & 66.9 & 39.6 & 37.7\\
\bottomrule
\end{tabular}
\end{table}

\subsection{Direct image encoder evaluation}\label{sec:exp:imgencoder}

\begin{table}[b]
  \newcolumntype{C}{>{\centering\arraybackslash}X}
  \newcolumntype{R}{>{\raggedleft\arraybackslash}X}
  \setlength{\tabcolsep}{0pt}
  \setlength{\extrarowheight}{5pt}
  \renewcommand{\arraystretch}{0.75}
  \centering
  \caption{Evaluations of the visual component in isolation (without the language model). We report fine-tuned classification accuracy on ImageNet, ImageNet-ReaL and ImageNet-V2; average zero-shot cross-modal retrieval recall@1 across 36 languages on XM3600; average 10-shot linear probe classification accuracy across 8 tasks. }\label{table:imagenet_transfer}
  \begin{tabularx}{\linewidth}{
  p{1.7cm}p{0.1cm}p{2.3cm}p{0.2cm}  
  Cp{0.1cm}Cp{0.1cm}Cp{0.1cm}Cp{0.2cm}  
  Cp{0.1cm}Cp{0.2cm}  
  C}  
    \toprule[1pt]
\multirow{3}{*}[4pt]{\bf{Model}} && \multirow{3}{*}[4pt]{\bf{Encoder}} &&
\multicolumn{7}{c}{\bf{ImageNet (fine-tuning)}} && \multicolumn{3}{c}{\bf{XM3600 (retrieval)}} &&
\bf{Probe} \\
     
    \cmidrule[0.5pt]{5-11} \cmidrule[0.5pt]{13-15} \cmidrule[0.5pt]{17-17}
    
 &&  &&
Res. && Val && ReaL && v2 &&
I$\rightarrow$T && T$\rightarrow$I &&
8 tasks \\

    \midrule
    
\NEWNAME && SigLIP ViT-G && 518px && 89.6 && 90.9 && 82.3 && 56.9 && 44.0 && 85.6 \\
 \midrule
PaLI-15B && Classif ViT-G    && 518px && 90.5 && 90.8 && 83.3 && - && - && 88.1 \\
PaLI-17B && Classif ViT-e    && 644px && 90.9 && 91.1 && 84.3 && 36.0 && 28.5 && 89.5 \\
PaLI-X   && Classif ViT-22B  && 756px && 89.2 && 91.0 && 83.7 && - && - && 89.9 \\
    \bottomrule[1pt]
  \end{tabularx}
\end{table}

Here, we aim to directly evaluate the learned image encoder (ViT-G model) without the surrounding language model, i.e.\ not the full \NEWNAME.
All results are summarized in Table~\ref{table:imagenet_transfer}.

First, we test image classification capabilities using the standard ImageNet \citep{russakovsky2014imagenet} benchmark and two of its most popular variants \citep{imagenet_real,recht2019imagenet}.
We fine-tune the unimodal image encoder from Stage~0 on ImageNet and compare it to fine-tuning of classification pretrained ViTs used in previous PaLI models.
The main comparison is to the classification (Classif) pretrained ViT-G/14 model from \citet{zhai2022scaling}.
The SigLIP slightly lags behind in terms of top-1 and v2 accuracy, but matches in terms of ReaL accuracy \citep{imagenet_real}, a metric which avoids measuring ``overfitting'' to ImageNet peculiarities.

Second, we report multilingual image-text retrieval results on the Crossmodal-3600 benchmark~\citep{Thapliyal2022Crossmodal3600AM}.
Since classification pretrained image encoders do not have this capability, we LiT-tune \citep{zhai2022lit} a text encoder for it on the multilingual WebLI dataset.
We also LiT-tune a new text encoder for the SigLIP image encoder in the exact same setting, to remove any confounders from the comparison.
The SigLIP ViT-G model clearly outperforms the classification pretrained larger ViT-e model.

Finally, we perform linear probing of the representation in the few-shot setting following \citet{vit,zhai2022scaling} across the 8 different classification tasks also used in \citet{zhai2022scaling} and report the average performance.
Here, we see that SigLIP lags behind, likely because the representation is not pretrained in a way that supports linear separability, as was recently uncovered by \citet{tschannen2023cappa}.

Taken together, these results paint a clear picture: the best and largest classification pretrained image encoders appear (slightly) better when evaluated on standard classification tasks, however they are significantly worse than SigLIP pretrained image encoders for vision-language tasks.

\section{Model Fairness, Biases, and Other Potential Issues}\label{sec:rai}

We follow the evaluation protocol of \citet{chen2023palix} to assess the model fairness, biases, and other potential issues. First, we use the MIAP~\citep{miap_aies} and FairFace~\citep{karkkainen2021fairface} datasets to generate captions and use the Perspective API~\citep{lees2022new} (threshold $>0.8$) to measure toxicity and profanity among other potential issues.  Table~\ref{table:toxicity_ff} (FairFace) and Table~\ref{table:toxicity_miap} (MIAP) summarize the results.  Slices with $<20$ examples are dropped from the analysis. Overall, we observe a low level of toxicity and profanity among others, across all slices. The results are comparable to those in PaLI-X~\citep{chen2023palix}.

\begin{table*}
\centering\small
\caption{RAI statistics for captions generated by \NEWNAME on FairFace~\citep{karkkainen2021fairface}.} 
\label{table:toxicity_ff}
\resizebox{\linewidth}{!}{%
\begin{tabular}{l|cc|ccc|ccc|c}
\toprule
 & \multicolumn{2}{c}{Perceived Gender} & \multicolumn{3}{c}{Ethnicity} & \multicolumn{3}{c}{Age Bucket} & \\
 &  Lowest & Highest & Lowest   & Median & Highest & Lowest & Median & Highest & \bf Overall\\
\midrule
\bf Toxicity &0.02\% & 0.05\% &  0.00\% & 0.00\% & 0.10\% & 0.00\% & 0.00\% & 0.07\% & \bf 0.04\%\\
\bf Profanity  & 0.00\% & 0.00\% & 0.00\% & 0.00\% & 0.00\% & 0.00\% & 0.00\% & 0.00\% & \bf 0.00\%\\
\bf Insult  & 0.04\% & 0.10\% & 0.00\% & 0.07\% & 0.14\% & 0.00\% & 0.06\% & 0.17\% & \bf 0.07\%\\
\bf Threat  & 0.10\% & 0.12\% & 0.00\% & 0.14\% & 0.20\% & 0.00\% & 0.00\% & 0.21\% & \bf 0.11\%\\
\bf Attack  & 0.00\% & 0.00\% & 0.00\% & 0.00\% & 0.00\% & 0.00\% & 0.00\% & 0.00\% & \bf 0.00\%\\
\bottomrule
\end{tabular}}
\end{table*}
\begin{table*}
\centering\small
\caption{RAI score statistics in the captions generated by \NEWNAME on MIAP~\citep{miap_aies}.} 
\label{table:toxicity_miap}
\resizebox{\linewidth}{!}{%
\begin{tabular}{l|cc|ccc|ccc|c}
\toprule
 & \multicolumn{2}{c}{Perceived Gender} &  \multicolumn{3}{c}{Age Bucket} & \multicolumn{3}{c}{Skin Tone} & \\
 &  Lowest & Highest & Lowest   & Median & Highest & Lowest   & Median & Highest & \bf Overall\\
\midrule
\bf Toxicity & 0.05\% &  0.10\% & 0.05\% & 0.24\% & 0.48\% & 0.00\% & 0.00\% & 0.26\% & \bf0.07\%\\
\bf Profanity & 0.10\% &  0.10\% & 0.00\% & 0.12\% & 0.24\% & 0.00\% & 0.00\% & 0.10\% & \bf0.10\%\\
\bf Insult & 0.10\% &  0.14\% & 0.00\% & 0.13\% & 0.17\% & 0.00\% & 0.00\% & 0.38\% & \bf0.13\%\\
\bf Threat & 0.34\% &  0.80\% & 0.30\% & 0.54\% & 1.68\% & 0.00\% & 0.59\% & 0.94\% & \bf0.65\%\\
\bf Identity Attack & 0.00\% &  0.00\% & 0.00\% & 0.00\% & 0.00\% & 0.00\% & 0.00\% & 0.00\% & \bf0.00\%\\
\bottomrule
\end{tabular}}
\end{table*}

\begin{figure}[t]
    \includegraphics[width=\columnwidth]{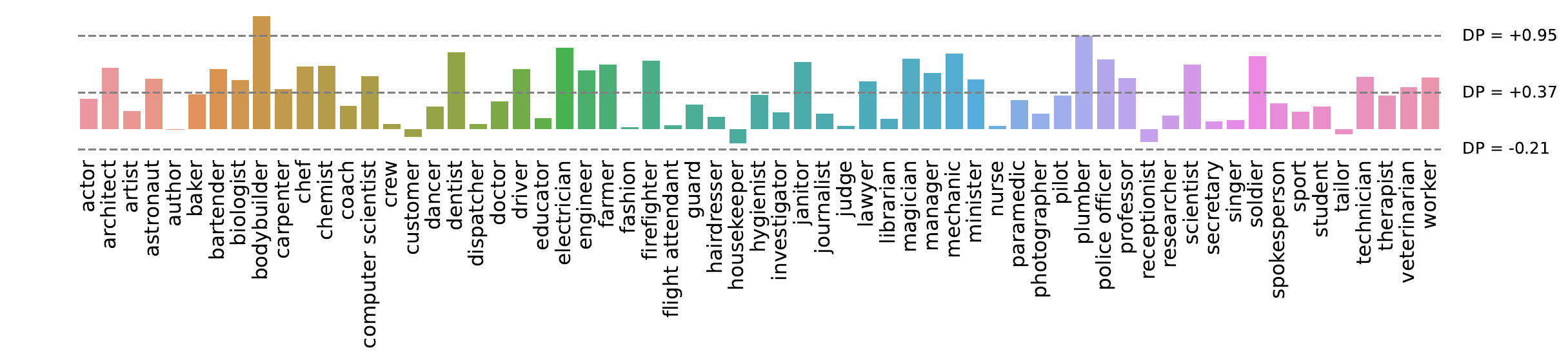}
    \caption{Level of demographic parity (DP) in \NEWNAME's output for CelebA images, comparing the average log-perplexity between females and males. Values close to zero indicate absence of bias. The dotted lines correspond to $\mu$ and $\mu\pm 2\sigma$. DP for all occupations falls within the 95\% confidence interval, except ``bodybuilder'' and ``plumber'' which the model tends to strongly associate with men.}
    \label{fig:representation_webli}
\end{figure}

Second, we examine the level of demographic parity \citep{dwork2012fairness} in the model itself. Following~\citep{chen2023palix}, we feed an image from the CelebA dataset \citep{liu2015faceattributes} into \NEWNAME with the chosen occupation as a prefix and record the average log-perplexity score of the model generation.
The demographic parity is the difference between average log-perplexity within the demographic groups. Figure~\ref{fig:representation_webli} summarizes the results: Similar to PaLI-X, \NEWNAME tends to assign a \emph{higher} log-perplexity score to women than men across most occupations with a mean difference of $\mu=0.37$. However, fewer occupations in \NEWNAME fall outside the interval $\mu\pm2\sigma$ compared to PaLI-X.

Third, we compare performance across all subgroups on a detection task using the MIAP dataset, following again~\citep{chen2023palix}. For images containing exactly a single individual, we query \NEWNAME with the question: ``Is there a person in this image?'' and evaluate the accuracy of its response. Table~\ref{table:miap_rai} summarizes the results. The error rate (false negatives) is very low across all subgroups.

\begin{table}[t]
\centering
\small
\caption{Detection error rate for ``person'' in \NEWNAME using the subset of the MIAP dataset~\citep{miap_aies} that contain exactly a single individual in the image. \NEWNAME maintains a low error rate across all subgroups. Skin tone follows the Monk Skin Tone Scale~\citep{Monk_2019}. Numbers inside square brackets correspond to the size of each bucket.
}
\label{table:miap_rai}
\resizebox{\linewidth}{!}{%
\begin{tabular}{lcccccccccc}
\toprule
\bf Skin Tone & {\bf1} {\scriptsize[2]} & {\bf2} {\scriptsize[871]} & {\bf3} {\scriptsize[3008]} & {\bf4} {\scriptsize[522]} & {\bf5} {\scriptsize[184]} & {\bf6} {\scriptsize[85]} & {\bf7} {\scriptsize[54]}& {\bf8} {\scriptsize[49]}& {\bf9} {\scriptsize[6]}& {\bf10} {\scriptsize[1]}
 \\[3pt]
& 0.00\% & 0.00\% & 0.17\% & 0.39\% & 0.00\% & 0.00\% & 0.00\% & 0.00\%  & 0.00\% & 0.00\% \\ \midrule
\bf Gender & \multicolumn{5}{c}{{\bf Predominantly Feminine} {\scriptsize[2437]}} & \multicolumn{5}{c}{{\bf Predominantly Masculine} {\scriptsize[3544]}} \\[3pt] 
  &       \multicolumn{5}{c}{0.41\%} & \multicolumn{5}{c}{1.78\%} \\
  \midrule
\bf Age Bucket & & \multicolumn{2}{c}{{\bf 0-2 yrs} {\scriptsize[17]}} & \multicolumn{2}{c}{{\bf 3-19 yrs} {\scriptsize[568]}} & \multicolumn{2}{c}{{\bf 20-59 yrs} {\scriptsize[4925]}} & \multicolumn{2}{c}{{\bf> 60 yrs} {\scriptsize[247]}} & \\[3pt]

 & & \multicolumn{2}{c}{0.00\%} & \multicolumn{2}{c}{0.18\%} & \multicolumn{2}{c}{1.30\%} & \multicolumn{2}{c}{0.82\%} & \\
\bottomrule
\end{tabular}}
\end{table}

For analysis of the WebLI dataset itself, such as the correlations between perceived gender and occupations, see \citet{chen2023palix}. 

\paragraph{Limitations.} The limitations of this work are very similar to those already presented in the literature. We refer to~\citep{chen2023palix}, which raises all limitations that apply to our work.

\section{Conclusion}\label{sec:conclusion}

In this paper, we took a closer look at the pretraining of the image encoder in large VLMs, specifically the PaLI type of model.
By performing controlled experiments on that part, for the first time, we clearly compare the two camps of classification pretraining and image-text (contrastive) pretraining, finding that the latter can lead to better and more efficient VLMs, especially for localization and text understanding tasks.
This is just \emph{one small aspect} of VLMs, and we hope this study and result spurs to further detailed investigations of the \emph{many other aspects} of training VLMs.

\section{Acknowledgements}
We would like to thank Bo Pang, Michael Tschannen, Soravit Changpinyo, AJ Piergiovanni, Marvin Ritter, Yi Tay, Paulina Pietrzyk, Matthias Minderer, André Susano Pinto, Mojtaba Seyedhosseini, Anelia Angelova, Neil Houlsby,
Yiming Gu, Daniel Hernandez Diaz, Chun-Liang Li, Guolong Su, Hao Zhang,
Chenxia Wu, Rui Lin, Ting Yu, Weina Ge, Shengyang Dai,
Tania Bedrax-Weiss, Rob Fergus, Jeremiah Harmsen, and Joelle Barral for helpful discussions, feedback, and support.
We thank the Google Research and Google DeepMind groups for providing a fruitful environment to conduct research.
\bibliography{egbib}
\bibliographystyle{iclr2024_conference}

\newpage

\appendix

\section{Additional results: Video Captioning and QA}
\label{appendix:video_results}
\subsection{Datasets \& Benchmarks}
Because not all videos in the benchmarks are available online at the time of experimentation when freshly collected the data, the effective numbers of videos is smaller than the public official splits in some cases. Table~\ref{table:video-data-wipeout} reports the details numbers for the subset used in our training and evaluation. We follows the same experimental settings used in PaLI-X \citep{chen2023palix}, including the dataset splits configuration and the evaluation metrics.
Please refer to PaLI-X \citep{chen2023palix} for more data-related details.

\begin{table}[ht]
\centering
\footnotesize
\resizebox{\linewidth}{!}{%
\begin{tabular}{llrrrrrrr}
\toprule
{}                 &       & MSR-VTT &  VATEX &  ANet-Cap &    SMIT &     M-V-QA &.       ANet-QA &  NExT-QA \\ \midrule
Original size      & valid.&     497 &   3000 &     17505 &   14604 &      12278 &          18000 &    5343 \\
                   & test  &    2990 &   6000 &     17031 &   3513  &      72821 &           8000 &    9178 \\ \midrule
Dataset size       & valid.&     325 &   2646 &     14566 &    8096 &       8160 &          10000 &    5343 \\
                   &  test &    2135 &   5242 &     14197 &    3513 &      52623 &           7040 &    9178 \\ \midrule
\% Remaining       & valid.&   65.39 &  88.20 &     83.21 &   100.00 &      66.46 &           - &  100.00 \\
                   &  test &   71.40 &  87.37 &     83.36 &   100.00 &      72.26 &          88.00 &  100.00 \\
\bottomrule
\end{tabular}}
\caption{As we freshly collect the data sets, the actual amount of training data is smaller than the public benchmarks, making the tasks more challenging. Except for NextQA and SMIT, there are more than 10\% of the videos missing in both training and evaluation.}

\label{table:video-data-wipeout}
\end{table}


\section{Additional results: Crossmodal-3600 retrieval}
\label{appendix:xm3600}

\begin{figure}[h]
    \includegraphics[width=\columnwidth]{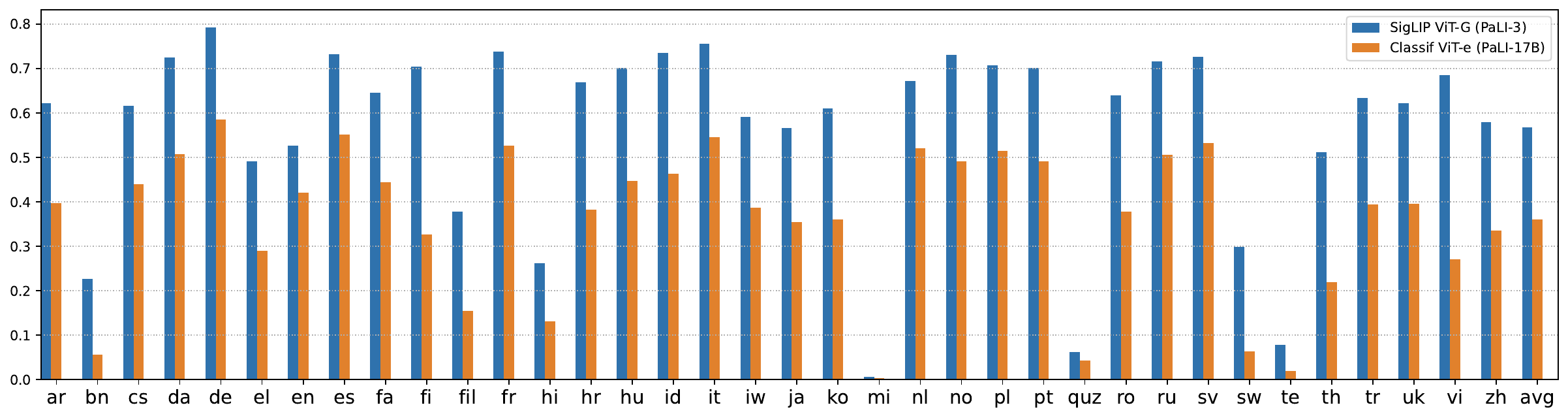}
    \caption{Image-to-text zero-shot retrieval recall@1 on crossmodal-3600 for SigLIP ViT-G and Classif ViT-e.}
    \label{fig:xm3600-i2t}
\end{figure}

\begin{figure}[h]
    \includegraphics[width=\columnwidth]{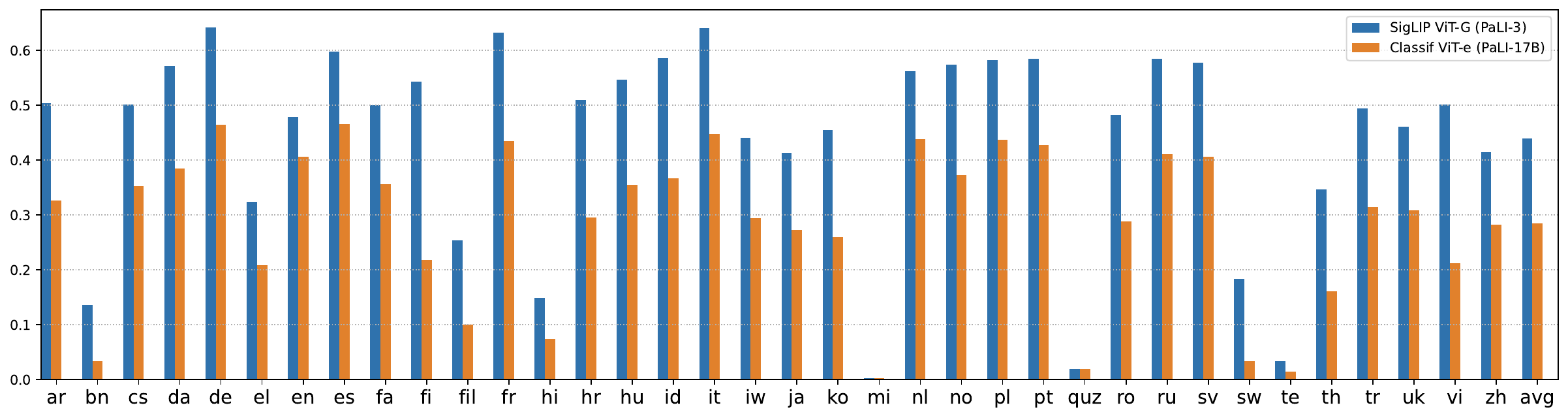}
    \caption{Text-to-image zero-shot retrieval recall@1 on crossmodal-3600 for SigLIP ViT-G and Classif ViT-e.}
    \label{fig:xm3600-t2i}
\end{figure}

We present more results of zero-shot image-text retrieval results (recall@1) on Crossmodal-3600~\citep{Thapliyal2022Crossmodal3600AM} in this section.
Detailed results across 36 languages for SigLIP ViT-G and Classif ViT-e are presented in Figure~\ref{fig:xm3600-i2t}, Figure~\ref{fig:xm3600-t2i} and Table~\ref{table:xm3600}.

\begin{table*}[ht]
\centering
\small
\begin{tabular}{lccrcc}
\toprule
\multirow{2}{*}{Language} & \multicolumn{2}{c}{Image-to-text} & & \multicolumn{2}{c}{Text-to-image} \\
\cmidrule{2-3} \cmidrule{5-6}
 & SigLIP ViT-G & Classif ViT-e && SigLIP ViT-G & Classif ViT-e \\
\midrule
ar & 62.22 & 39.69 && 50.33 & 32.60 \\
bn & 22.67 & 5.67 && 13.61 & 3.31 \\
cs & 61.69 & 44.03 && 50.10 & 35.24 \\
da & 72.47 & 50.75 && 57.16 & 38.48 \\
de & 79.28 & 58.53 && 64.20 & 46.50 \\
el & 49.11 & 29.03 && 32.36 & 20.92 \\
en & 52.64 & 42.11 && 47.89 & 40.63 \\
es & 73.31 & 55.22 && 59.77 & 46.55 \\
fa & 64.56 & 44.50 && 50.09 & 35.58 \\
fi & 70.39 & 32.64 && 54.34 & 21.80 \\
fil & 37.89 & 15.53 && 25.40 & 10.04 \\
fr & 73.81 & 52.61 && 63.28 & 43.47 \\
hi & 26.22 & 13.14 && 14.94 & 7.42 \\
hr & 66.89 & 38.31 && 51.03 & 29.55 \\
hu & 70.22 & 44.67 && 54.63 & 35.49 \\
id & 73.53 & 46.33 && 58.62 & 36.75 \\
it & 75.56 & 54.53 && 64.14 & 44.76 \\
iw & 59.14 & 38.67 && 44.10 & 29.39 \\
ja & 56.69 & 35.47 && 41.31 & 27.24 \\
ko & 61.03 & 36.11 && 45.50 & 25.95 \\
mi & 0.64 & 0.33 && 0.30 & 0.22 \\
nl & 67.25 & 52.14 && 56.23 & 43.79 \\
no & 73.03 & 49.17 && 57.40 & 37.35 \\
pl & 70.69 & 51.42 && 58.24 & 43.72 \\
pt & 70.22 & 49.19 && 58.47 & 42.73 \\
quz & 6.28 & 4.31 && 1.89 & 1.90 \\
ro & 63.92 & 37.75 && 48.20 & 28.82 \\
ru & 71.69 & 50.64 && 58.51 & 41.11 \\
sv & 72.69 & 53.22 && 57.76 & 40.66 \\
sw & 29.86 & 6.42 && 18.31 & 3.41 \\
te & 7.81 & 1.92 && 3.39 & 1.42 \\
th & 51.14 & 22.00 && 34.65 & 16.06 \\
tr & 63.36 & 39.50 && 49.41 & 31.47 \\
uk & 62.19 & 39.53 && 46.15 & 30.81 \\
vi & 68.50 & 27.08 && 50.14 & 21.28 \\
zh & 57.92 & 33.61 && 41.51 & 28.24 \\
\midrule
avg & 56.85 & 35.99 && 43.98 & 28.46 \\
\bottomrule
\end{tabular}
\caption{Crossmodal-3600 zero-shot retrieval recall@1 (\%) for SigLIP ViT-G (2B params) and Classif ViT-e (4B params).
SigLIP ViT-G is significantly better than Classif ViT-e across all the languages.
SigLIP improves from 28.5\% to 44.0\% on average for the text-to-image retrieval task.}
\label{table:xm3600}
\end{table*}

\end{document}